\documentclass{IEEEoj}
\usepackage{cite}
\usepackage{amsmath,amssymb,amsfonts}
\usepackage{algorithmic}
\usepackage{graphicx,color}
\usepackage{textcomp}
\usepackage[T1]{fontenc}
\usepackage[caption=false,font=normalsize,labelfont=sf,textfont=sf]{subfig}
\usepackage{multirow}
\usepackage{float}
\usepackage{placeins}
\usepackage{arydshln}

\def\BibTeX{{\rm B\kern-.05em{\sc i\kern-.025em b}\kern-.08em
    T\kern-.1667em\lower.7ex\hbox{E}\kern-.125emX}}
\AtBeginDocument{\definecolor{ojcolor}{cmyk}{0.93,0.59,0.15,0.02}}
\def\OJlogo{\vspace{-14pt}\includegraphics[height=28pt]{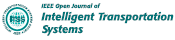}}
\begin{document}
\receiveddate{04 July, 2025}
\reviseddate{XX Month, XXXX}
\accepteddate{XX Month, XXXX}
\publisheddate{XX Month, XXXX}
\currentdate{XX Month, XXXX}
\doiinfo{OJITS.2022.1234567}

\newcommand{\mytitle}{CSNR and JMIM Based Spectral Band Selection for Reducing Metamerism in Urban Driving}

\title{\mytitle}

\author{Jiarong Li\textsuperscript{1,2}, Imad Ali Shah\textsuperscript{1,2}, Diarmaid Geever\textsuperscript{1,2}, Fiachra Collins\textsuperscript{3}, Enda Ward\textsuperscript{3}, Martin Glavin\textsuperscript{1,2}, Edward Jones\textsuperscript{1,2}, Brian Deegan\textsuperscript{1,2}}
\affil{School of Engineering, University of Galway, Ireland}
\affil{Ryan Institute, University of Galway, Ireland}
\affil{Valeo Vision Systems, Tuam, Ireland}

% \corresp{CORRESPONDING AUTHOR: J. Li (e-mail: j.li11@universityofgalway.ie).}
% \authornote{This work was produced by the Connaught Automotive Research (CAR) group at the University of Galway and was supported, in part, by Taighde Éireann – Research Ireland grants 13/RC/2094 P2 and 18/SP/5942, and co-funded under the European Regional Development Fund through the Southern and Eastern Regional Operational Programme to Lero - the Research Ireland Centre for Software (www.lero.ie), and by Valeo Vision Systems.}

\authornote{This work has been submitted to the IEEE for possible publication. Copyright may be transferred without notice, after which this version may
no longer be accessible.}

\markboth{\mytitle}{Li \textit{et al.}}

\begin{abstract}
Protecting Vulnerable Road Users (VRU) is a critical safety challenge for automotive perception systems, particularly under visual ambiguity caused by metamerism, a phenomenon where distinct materials appear similar in RGB imagery. This work investigates hyperspectral imaging (HSI) to overcome this limitation by capturing unique material signatures beyond the visible spectrum, especially in the Near-Infrared (NIR). To manage the inherent high-dimensionality of HSI data, we propose a band selection strategy that integrates information theory techniques (joint mutual information maximization, correlation analysis) with a novel application of an image quality metric (contrast signal-to-noise ratio) to identify the most spectrally informative bands. Using the Hyperspectral City V2 (H-City) dataset, we identify three informative bands (497 nm, 607 nm, and 895 nm, $\pm$27 nm) and reconstruct pseudo-color images for comparison with co-registered RGB. Quantitative results demonstrate increased dissimilarity and perceptual separability of VRU from the background. The selected HSI bands yield improvements of 70.24\%, 528.46\%, 1206.83\%, and 246.62\% for dissimilarity (Euclidean, SAM, $T^2$) and perception (CIE $\Delta E$) metrics, consistently outperforming RGB and confirming a marked reduction in metameric confusion. By providing a spectrally optimized input, our method enhances VRU separability, establishing a robust foundation for downstream perception tasks in Advanced Driver Assistance Systems (ADAS) and Autonomous Driving (AD), ultimately contributing to improved road safety.
\end{abstract}



\maketitle

\section{INTRODUCTION}
\label{intro}
\IEEEPARstart{A}{utonomous} Driving (AD) and Advanced Driver Assistance Systems (ADAS) promise a safer future, but their ability to reliably detect Vulnerable Road Users (VRU) is fundamentally limited by their lack of sensitivity to non-visible wavelengths, creating potentially unseen threats. As the most physically exposed road users, such as pedestrians and cyclists, VRU is highly susceptible to severe outcomes in traffic incidents, making their timely and accurate detection a critical safety requirement. This challenge is intensified in complex real-world scenarios with diverse lighting conditions, cluttered backgrounds, and partial occlusions.

\begin{figure*}
\centerline{\includegraphics[width=\textwidth]{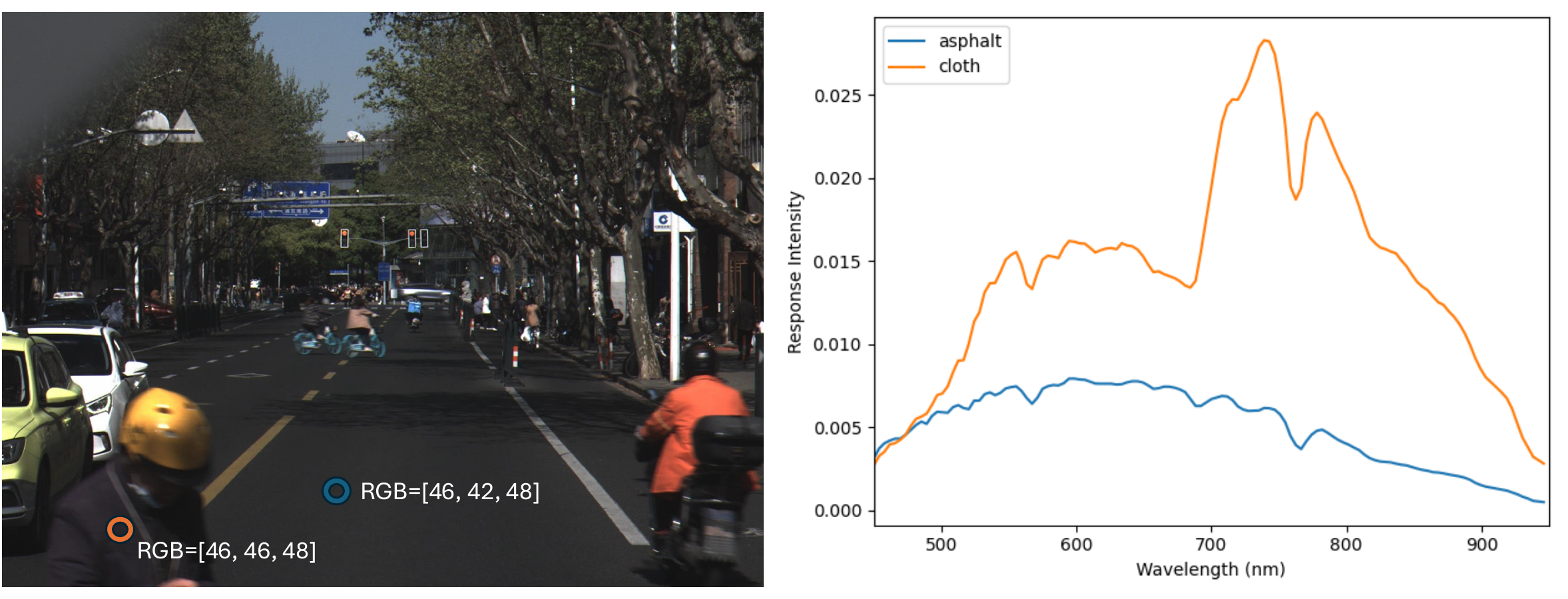}}
\caption{Illustration of metamerism and its impact on visual ambiguity, using the Hyperspectral City V2~\cite{shen4560035urban} (H-City) dataset. The left panel shows a standard RGB image where pixels within the orange (cloth) and blue (asphalt) circles appear nearly identical in color, making them visually indistinguishable. The panel on the right hand side plots the corresponding hyperspectral reflectance curves for these regions, revealing significantly distinct spectral signatures, highlighting how metamerism can mask material differences when relying solely on RGB sensor data.
\label{fig_metamerism}}
\end{figure*}

% Explain the problem of metamerism 
A major factor exacerbating these difficulties is metamerism, where the colors of clothing or objects may match the background under specific illumination but differ under another~\cite{akbarinia2018color}, as illustrated in Fig.~\ref{fig_metamerism}. Metamerism directly causes visual confusion, reducing the contrast and distinctiveness needed for reliable detection by vision-based systems operating under conditions where a metameric match occurs.

% limitations of RGB-based vision systems
Modern automotive systems employ a suite of complementary sensors, including RGB cameras, LiDAR, and radar, to continuously scan and monitor the environment~\cite{van2018autonomous}. LiDAR offers precise 3D structure through point clouds but lacks color and texture information, whereas radar performs reliably in adverse weather conditions for robust speed and range detection~\cite{campbell2018sensor}. RGB cameras, on the other hand, provide rich visual features, essential for detailed object recognition and lane keeping~\cite{campbell2018sensor}. However, designed to mimic human vision, these cameras are inherently susceptible to the same visual challenges, struggling with detection in low-contrast situations, variable lighting, and when metameric effects render important objects indistinguishable in the visible spectrum~\cite{thornton1998strong}. These limitations create a bottleneck for achieving robust and reliable VRU detection.

% Introduce hyperspectral imaging
Hyperspectral sensors offer a compelling solution by capturing light across numerous narrow bands spanning a broader range of the electromagnetic spectrum. These sensors generate detailed spectral signatures for each spatial pixel, enabling finer material differentiation and detection of subtle variations invisible to standard RGB cameras~\cite{lu2020recent}. This enhanced discriminative capability is particularly promising for resolving metameric ambiguity and improving object visibility under challenging conditions.

% State the HSI Challenge
Despite these advantages, the real-time use of Hyperspectral Imaging (HSI) data in automotive applications poses significant challenges due to high data dimensionality and redundancy~\cite{arneal2015spectral}. This increased spectral resolution comes at the cost of higher computational complexity and reduced photon capture per band, leading to lower Signal-to-Noise ratios (SNR) and degraded low-light performance~\cite{zhang2023snapshot}. Effectively mitigating these challenges necessitates an efficient approach to select and represent the most relevant spectral information from the HSI data for reliable VRU identification using a limited subset of wavebands. Thus, to leverage HSI data effectively for real-time, resource-constrained applications, efficient band selection is crucial to isolate the most informative wavelengths.

% Contribution/Solution
This paper addresses these challenges by focusing on enhancing VRU-background separability, a critical factor for improving road safety in automotive perception systems. The objective is to identify spectral bands that maximize visual and computational discriminability between VRU and the environment. To achieve this, the paper makes the following key contributions:

\begin{enumerate}
\item This paper proposes a spectral band selection strategy that integrates Joint Mutual Information Maximization
(JMIM)~\cite{bennasar2015feature}, inter-band correlation analysis~\cite{de2016comparing}, Contrast Signal-to-Noise Ratio (CSNR)~\cite{klein2023evaluation} to identify informative bands while preserving visual quality.
\item This paper investigates the role of the Near-Infrared (NIR) region in mitigating metameric ambiguity and enhancing VRU-background separability.
\item This paper validates the approach using the Hyperspectral City V2~\cite{shen4560035urban} (H-City) dataset, quantitatively demonstrating substantial improvements over RGB images using dissimilarity and perceptual-based metrics.

\end{enumerate}

The remainder of this paper is organized as follows. Section~\ref{related_work} reviews related work in automotive perception, VRU detection, and HSI processing. Section~\ref{methodology} details the proposed band selection framework. Section~\ref{experiment} describes the experimental setup and evaluation protocol. Section~\ref{results} presents our findings and quantitative comparisons. Section~\ref{discussion} offers interpretation and limitations, and Section~\ref{conclusion} concludes the paper with directions for future research.

\begin{table}
\caption{Sensor Modalities Used For Automotive Perception}
\label{table1}
\setlength{\tabcolsep}{2.5pt} %3pt
\begin{tabular}{>{\raggedright\arraybackslash}p{25pt}
                >{\raggedright\arraybackslash}p{105pt}
                >{\raggedright\arraybackslash}p{105pt}} %{|p{25pt}|p{75pt}|p{115pt}|}
\hline
\textbf{Sensors} & 
\textbf{Advantages} & 
\textbf{Limitations} \\
\hline
RGB & 
High spatial resolution, rich color and texture information, passive sensing~\cite{silva2024vulnerable}, low costs and power consumption~\cite{sivaraman2013looking} & 
Sensitivity to lighting conditions and weather conditions, lack of speed and depth information~\cite{zhang2023perception} \\
\hline
LiDAR& 
Accurate depth perception, day/night operation~\cite{kumar2023object} & Performance degrades in adverse weather, high costs~\cite{zhang2023perception}
 \\
\hline
Radar& 
Speed and distance measurement, robust in adverse weather~\cite{zhang2023perception} & Low resolution, poor pedestrian detection~\cite{ullmann2023survey}  \\

\hline

\end{tabular}
\label{tab1}
\end{table}

\section{RELATED WORK}
\label{related_work}

\subsection{VRU Detection Challenges in Automotive Perception}
Reliable VRU detection is critical for ADAS safety, but current automotive perception systems face fundamental limitations. Multi-modal sensor fusion combining RGB cameras, LiDAR, and radar exhibits complementary weaknesses that constrain VRU detection under challenging conditions~\cite{silva2024vulnerable,zhang2023perception}.

RGB cameras provide rich visual information but suffer severe degradation under variable illumination and adverse weather~\cite{van2018autonomous}. LiDAR offers accurate depth perception but experiences signal attenuation in precipitation~\cite{kumar2023object}, while radar lacks spatial resolution for reliable VRU shape classification~\cite{ullmann2023survey}. Deep learning approaches, including single-stage detectors like YOLO~\cite{redmon2016you} and two-stage methods like Faster R-CNN~\cite{ren2015faster}, have improved detection performance but remain fundamentally constrained by input data quality.

A critical challenge is metamerism, where VRU clothing appears visually similar to background materials under specific illumination despite having different spectral properties~\cite{akbarinia2018color,thornton1998strong}. This phenomenon reduces visual contrast essential for RGB-based detection, creating scenarios where VRUs become effectively invisible to conventional perception systems. When environmental conditions simultaneously degrade multiple sensor modalities, fusion cannot recover discriminative information that was never captured, necessitating alternative sensing approaches.

\subsection{HSI for ADAS/AD Applications}
HSI addresses RGB limitations by capturing hundreds of narrow spectral bands across extended electromagnetic spectra, generating unique material-specific signatures that enable discrimination independent of visual appearance~\cite{lu2020recent}. This capability offers significant advantages for resolving metameric ambiguity in VRU detection.

Although automotive HSI research is in early stages~\cite{shah2025multi}, it has already demonstrated potential applications for ADAS/AD. Winkens et al.~\cite{winkens2017hyperspectral} showed terrain classification capabilities in off-road scenarios, while Gutierrez et al.~\cite{gutierrez2023chip} achieved improved object segmentation by exploiting material reflectance properties invisible to RGB sensors. NIR-specific studies show particular promise: Broggi et al.~\cite{broggi2006pedestrian} demonstrated enhanced pedestrian detection using NIR cameras in moving vehicles, and Nataprawira et al.~\cite{nataprawira2021pedestrian} achieved improved accuracy in low-light scenarios through VIS-NIR fusion.

The Near-Infrared region (700-1000 nm) exhibits unique advantages for ADAS applications: enhanced atmospheric penetration, maintained solar illumination, and distinct material reflectance characteristics that differ from visible appearance~\cite{hertel2009low,weikl2022potentials}. Research by Herweg et al.~\cite{herweg2012hyperspectral,herweg2013separability} demonstrated HSI's ability to distinguish individuals and clothing materials under varying conditions, providing enhanced discrimination where visible contrast is insufficient.

However, HSI deployment faces significant challenges. High-dimensional data creates computational burdens for real-time processing~\cite{yuan2014hyperspectral}, while increased spectral resolution reduces photon capture per band, degrading low-light performance~\cite{zhang2023snapshot}. These constraints necessitate efficient band selection strategies for automotive implementation.

\subsection{Hyperspectral Band Selection Methods}
Band selection addresses HSI's dimensionality challenge by identifying optimal spectral subsets that retain discriminative information while reducing computational load~\cite{Sawant2020}. Filter-based methods are particularly suitable for ADAS applications due to their computational efficiency and classifier independence~\cite{Saeys2007}.

Information-theoretic approaches dominate current band selection research. Mutual Information (MI) measures statistical dependence between bands and target classes~\cite{kraskov2004estimating}. Joint Mutual Information Maximization (JMIM) extends this concept by selecting bands that maximize information gain while minimizing redundancy~\cite{bennasar2015feature}.

Correlation analysis identifies redundant bands through inter-band correlation assessment. Pearson correlation coefficients quantify linear relationships, enabling the removal of highly correlated bands that provide similar information~\cite{ma2024socf}. Advanced methods, such as Spectral Correlation-based Feature Selection (SOCF), combine correlation with spectral similarity measures.

Distance-based methods utilize spectral similarity metrics such as Spectral Angle Mapper (SAM) to identify representative bands through clustering approaches~\cite{9829585}. These methods ensure spectral diversity in selected subsets by choosing representatives from spectrally distinct clusters.

Recent hybrid approaches combine multiple criteria for improved performance. Relief-F considers feature interactions and class separability~\cite{Guyon2006}, while Fast Correlation-Based Filter (FCBF) integrates correlation analysis with information theory~\cite{JMLR:v13:brown12a}. However, existing methods inadequately address automotive-specific requirements, including real-time constraints and metameric disambiguation.

\subsection{Evaluation Metrics and Research Gaps}

Effective band selection evaluation requires metrics balancing dimensionality reduction with information preservation for ADAS applications.

Separability metrics quantify discrimination capability between VRU and background regions using statistical distance measures. Euclidean distance provides basic separability assessment, while Mahalanobis distance accounts for class covariance structure~\cite{webb2003statistical}. Bhattacharyya distance measures separability between probability distributions, particularly relevant for overlapping classes in VRU scenarios~\cite{guorong1996bhattacharyya}.

Perceptual quality metrics assess visual information preservation crucial for ADAS applications. CSNR measures local contrast preservation~\cite{watson2000visual}, ensuring selected bands maintain visual interpretability for human operators and downstream processing algorithms.

Performance metrics directly evaluate task effectiveness through classification accuracy, precision, and recall for VRU detection applications~\cite{MEDJAHED2018413}. Information-theoretic metrics like entropy provide classifier-independent assessment of band informativeness~\cite{Peter2004}.

Critical research gaps exist in automotive HSI band selection. Current filter methods inadequately balance redundancy reduction with discriminatory information preservation, particularly for safety-critical VRU detection where subtle material differences are crucial~\cite{theng2024feature}. Existing approaches focus primarily on remote sensing applications with insufficient consideration of automotive constraints: real-time processing requirements, diverse environmental conditions, and metameric effects.

Limited availability of automotive HSI datasets constrains validation across representative scenarios. Current evaluation frameworks lack integration of perceptual quality assessment with information-theoretic metrics, essential for ADAS applications requiring both computational efficiency and visual interpretability.

This work addresses these limitations through a novel filter-based methodology integrating JMIM, correlation analysis, and CSNR metrics, specifically optimized for automotive VRU detection and metameric disambiguation using NIR spectral information.

\section{METHODOLOGY}
\label{methodology}
Our band selection strategy is designed to identify a spectrally diverse set of bands that maximizes statistical information and perceptual contrast for critical VRU separability in ADAS/AD safety tasks. To achieve this, we developed a multi-criteria framework that evaluates each band based on information-theoretic value, spectral redundancy, and perceptual contrast. This approach avoids the pitfalls of relying on a single metric and ensures a robust, balanced selection. The process is detailed below.

\subsection{CRITERION 1: INFORMATION MAXIMIZATION VIA JMIM}
To ensure the selected bands are rich in class-specific information, we employ the Joint Mutual Information Maximization (JMIM) ~\cite{bennasar2015feature}. In the context of HSI, JMIM is a powerful feature selection technique that iteratively selects bands that maximize the information they jointly share with a target classification, while minimizing redundancy with already selected bands~\cite{brown2012conditional}. This is crucial for identifying bands that capture the most discriminative spectral signatures.

The JMIM selection process is defined as:
$$
f_{JMIM}=\operatorname*{arg\,max}_{f_{i}\in{F-S}}(min_{f_s\in{S}}(I(f_i,f_s;C))),
$$
where $f_i$ is a candidate band from the set of all bands $F$ not yet in the selected set $S$, and $I(f_i,f_s;C)$ is the joint mutual information between the candidate band, an already selected band $f_s$, and the target class $C$. In our framework, we use JMIM to generate a top-ranked list of candidate bands that are highly informative for the perception task.

\subsection{CRITERION 2: REDUNDANCY ANALYSIS VIA INTER-BAND CORRELATION}
To complement JMIM and explicitly minimize data redundancy, we perform an inter-band correlation analysis. High correlation between bands indicates overlapping information, which increases computational load without adding new perceptual value~\cite{manolakis2008spectral}. 

We use the Pearson correlation coefficient $\rho_{ij}$~\cite{de2016comparing}, to quantify the linear relationship between any two bands, $x_i$ and $x_j$. The correlation coefficient is calculated as:

$$
\rho_{ij} = \frac{Cov(x_{i}x_{j})}{\sqrt{Var(x_i)Var(x_j)}} ,\qquad -1\leq\rho_{ij}\leq 1.
$$

To represent pairwise correlations across the entire set of bands, we calculate the overall correlation matrix $P$. This matrix is derived from the covariance matrix $\Sigma$, normalized by the standard deviations of the individual bands $\sigma_{ii}$. The element $P_{ij}$ corresponds to $\rho_{ij}$, and we define the matrix as:

$$
P = D(\frac{1}{\sqrt{\sigma_{ii}}})\Sigma D(\frac{1}{\sqrt{\sigma_{ii}}}),
$$
where $\Sigma$ represents the covariance matrix of the HSI data, $\sigma_{ii}$ are its diagonal elements, representing the variance of the band $i$, and $D(\cdot)$ denotes a diagonal matrix formed from the specified elements.

This correlation matrix serves two roles in our framework.  During selection, it is first used to penalize candidate bands that are highly correlated with other high-ranking candidates. It is also applied in a final refinement step to ensure spectral diversity across the final band set.

\subsection{CRITERION 3: PERCEPTUAL CONTRAST ESTIMATION VIA CSNR}
A band set can be statistically optimal but still produce poor visual results. To ensure the selected bands generate high-contrast imagery, we evaluate them using the CSNR metric. CSNR measures a sensor's ability to distinguish between two regions by assessing the ratio of the mean contrast $\Bar{c}$ to its standard deviation $\sigma_c$ across multiple samples: 
$$
CSNR = \frac{\Bar{c}}{\sigma_c},
$$

where we calculate $\Bar{c}$ using the Michelson contrast~\cite{peli1990contrast}. A high CSNR indicates that a band consistently produces a strong, stable contrast signal for key material pairs, such as clothing against asphalt, making it robust for real-world detection~\cite{klein2023evaluation}. In our framework, we calculate a CSNR matrix to estimate the contrast potential for all bands, guiding us toward selections that promise strong visual separability.

\subsection{MULTI-CRITERIA INFORMED SELECTION PROCESS}
Our final selection process synthesizes these three criteria in a guided, multi-step process:

\subsubsection{Candidate Identification}
We first identify a pool of high-potential candidates based on the top-ranking bands from the JMIM analysis and the CSNR matrix, along with the low-correlation bands from the correlation matrix.

\subsubsection{Informed Selection}
Instead of performing a global score integration across all bands, the initial band subset is selected by analyzing the individual metric results. Although it is theoretically desired to compute metrics for all possible band combinations, the significant computational expense of JMIM prohibited integrating a full matrix score across the high-dimensional HSI cube. To navigate this challenge and still leverage the strengths of multiple criteria, we focus on identifying bands that demonstrate strong evidence across the criteria simultaneously, including consistently appearing among the top JMIM candidates, exhibiting low correlation with other potential selections, and showing high estimated CSNR values for key material contrasts relevant to ADAS/AD perception.

\subsubsection{Spectral Diversity Check}
The initial selection is refined to ensure maximal spectral diversity. If two selected bands are spectrally adjacent and highly correlated, one is retained and the other is replaced with the next-best qualifying candidate from the pool to maximize the uniqueness of the information in the final set.

This integrated process guarantees the final band set is not merely a product of one metric but represents a consensus across statistical, redundancy, and perceptual criteria, making it fundamentally more robust.

\section{EXPERIMENTAL SETUP}
\label{experiment}
This section specifies the data, baselines, and evaluation strategy used to validate our band selection methodology. The primary goal is to demonstrate enhanced separability for VRU in complex urban scenes.

\subsection{HSI DATASET}
We utilized the H-City dataset from the available HSI datasets, as it is the only one that provides co-registered RGB and HSI image pairs~\cite{shah2024hyperspectral}, along with high spatial (1889 × 1422 pixels) and spectral resolution (450–950 nm, 128 bands). Its superior spectral resolution and greater diversity of urban scenarios, lighting conditions, and VRU instances make it ideal for this research.

To develop a robust band selection model, we curated a representative data subset from H-City. This involved manually selecting 30 distinct material patches under consistent daylight conditions. Each 10\texttimes 10 pixel patch was chosen to ensure material uniformity, with a focus on materials critical to VRU separability. These included VRU materials, such as various clothing fabrics; background materials, such as asphalt and concrete; and common urban elements, such as vehicle tires, road markings, and vegetation.

The material composition of these patches was validated by comparing their spectral signatures to established references in the NASA ECOSTRESS Spectral Library~\cite{meerdink2019ecostress, baldridge2009aster}. This curated set of material signatures formed the foundation for applying our band selection methodology.

\begin{figure*}[!t]
\centering
\subfloat[]{\includegraphics[width=2.7in]{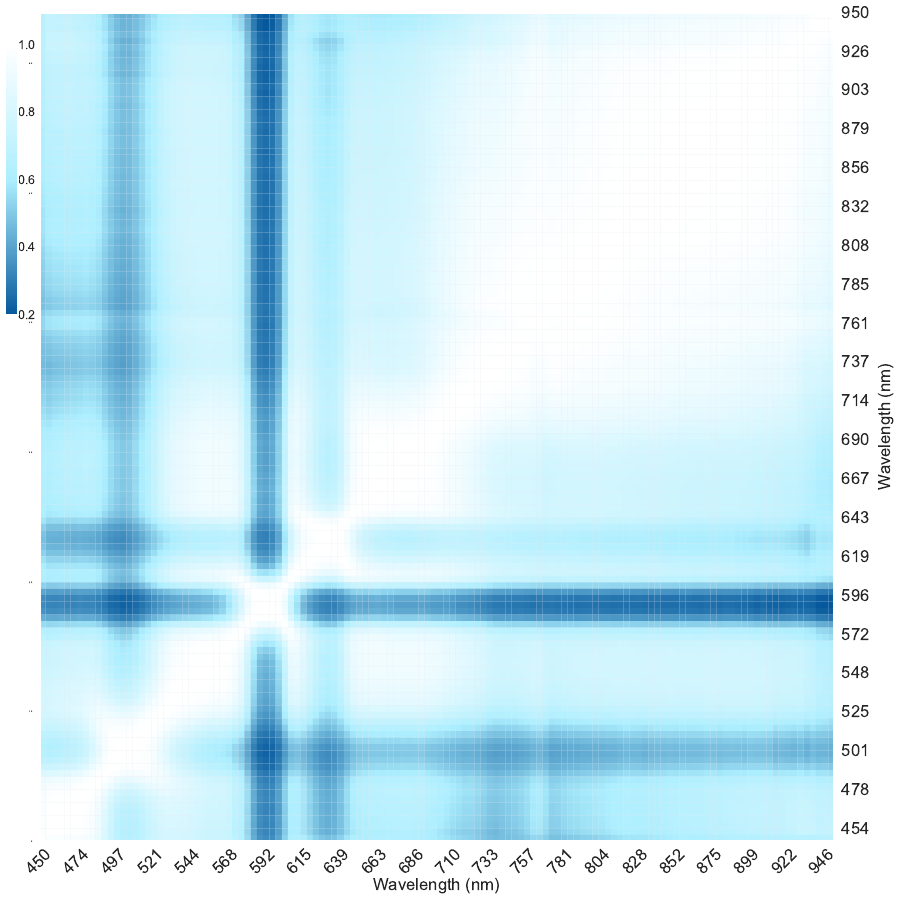}%
\label{fig_corr}}
\hfil
\subfloat[]{\includegraphics[width=4in]{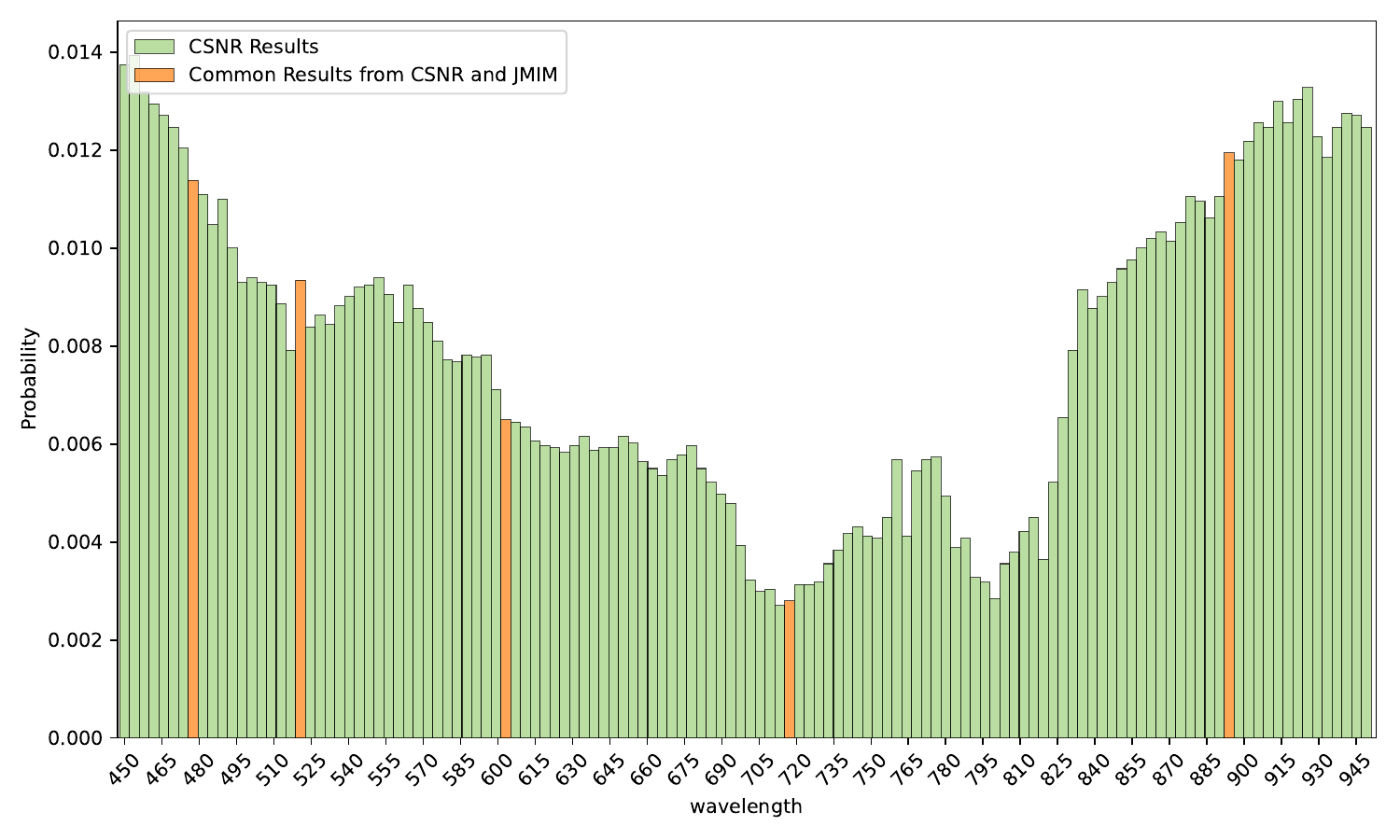}%
\label{fig_csnr}}
\caption{Individual Metric Analyses for Initial Band Assessment from the H-City Dataset.
(a) Inter-Band Correlation Heatmap. The heatmap illustrates pairwise correlation coefficients across all spectral bands, ranging from 0.2 to 1.0. White clusters indicate highly correlated bands, and deep blue clusters indicate weakly correlated bands.
(b) CSNR Probability Distribution. This plot highlights the probability of bands exhibiting high CSNR values, indicating their potential perceptual quality. Initial JMIM candidate bands are marked in orange.}
\label{fig_indiv}
\end{figure*}

\begin{figure}
\centerline{\includegraphics[width=0.485\textwidth]{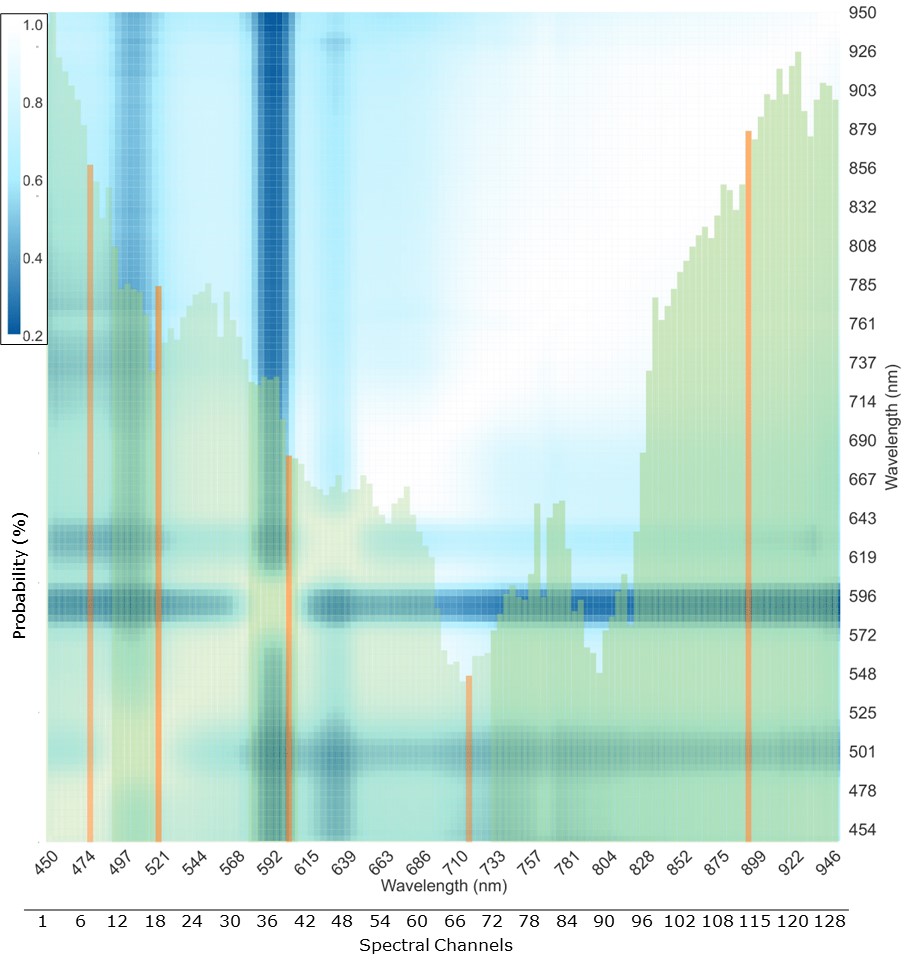}}
\caption{Integrated Multi-Metric Analysis for Optimal Band Selection. The heatmap shows inter-band correlation coefficients across the spectral range from 450 nm to 950 nm (128 channels), where deep blue represents low correlation (0.2) and white indicates high correlation (1.0). The probability distribution of bands with high CSNR values is overlaid. Orange vertical bars mark the initial five candidate bands identified by JMIM. This combined visualization highlights consensus regions exhibiting low correlation, high CSNR probability, and strong JMIM scores, guiding the final selection of the 497 nm, 607 nm, and 895 nm wavelengths.
\label{fig1}}
\end{figure}

\subsection{EVALUATION BASELINES}
We assessed the performance of our selected bands against two well-defined baselines.

The primary baseline is the original RGB images from the H-City dataset, representing the current industry standard for visual perception. The second baseline is derived from our band selection methodology. After identifying an optimal band set of 3 spectral bands, we reconstructed HSI data for the entire dataset. This achieves a reduction of over $97\%$ in data volume (from 128 to 3 bands) and serves as the foundation for our quantitative analysis and pseudo-color composites.

The core of our evaluation is a direct comparison of class separability and visual discriminability between the standard RGB images and our 3-band HSI reconstructions.

\subsection{EVALUATION PROTOCOL}
Our evaluation protocol is designed to measure the inherent discriminatory power of the spectral bands. Given the high spatial and spectral resolution and large data volume of HSI data, we focus on patch-based analysis.

The protocol involves comparing the standard RGB baseline against our reconstructed HSI representation using a suite of quantitative metrics. We assess performance with class dissimilarity measures, including Euclidean Distance $D_2$, Spectral Angle Mapper (SAM)~\cite{9829585}, and the Multivariate $T^2$ test~\cite{murphy1987selecting}, to quantify mathematical separability. Additionally, we use the visual perception metric CIE $\Delta E$~\cite{choudhury1996evaluation} to evaluate perceptual differences and the potential for metameric confusion. The detailed definitions and applications of these metrics are presented alongside their corresponding results in Section~\ref{results}.

\section{RESULTS}
\label{results}
This section presents the effectiveness of our optimized spectral band strategy in improving the perception of VRU. It also assesses the perceptual performance on other common traffic elements, particularly traffic lights and road markings. We first detail the principled process for identifying an optimized band set, then evaluate these selected bands through qualitative visual assessments and quantitative metrics.

\subsection{SPECTRAL BANDS SELECTION}
\label{results:bands}
Our band selection methodology was applied to the curated H-City data subset to identify a minimal yet powerful set of bands. The process involved a multi-criteria analysis to find the optimal balance between statistical information (JMIM), spectral uniqueness (inter-band correlation), and perceptual contrast (CSNR).

\subsubsection{Step 1: Initial Candidate Generation}
The process began by using JMIM to identify the five most informative bands out of 128 channels in H-City for distinguishing key materials in our dataset. This yielded an initial ranked candidate set of channels: 114 (895 nm), 7 (477 nm), 40 (607 nm), 18 (520 nm), and 68 (716 nm), as illustrated in Fig.~\ref{fig_indiv}--\ref{fig1}. The wavelengths are rounded to the nearest number. These bands served as the starting point for multi-criteria refinement. Concurrently, we used inter-band correlation to identify spectral redundancy, and CSNR probability distributions to highlight bands with high potential perceptual quality. Figure~\ref{fig_indiv} presents these initial individual metric analyses.

\subsubsection{Step 2: Multi-Criteria Evaluation and Refinement}
Next, we evaluated this initial set of candidates using an integrated visualization (Fig.~\ref{fig1}) that overlays the inter-band correlation heatmap with the CSNR probability distribution. This allowed us to assess each band's trade-offs holistically.

The 895 nm (NIR) and 607 nm (Red) bands were immediately confirmed as strong selections. Both were not only highly ranked by JMIM but also resided in regions of very low correlation ($<0.3$) and demonstrated high CSNR probability, indicating they provide unique, high-contrast information.

The 716 nm band, despite its high JMIM rank, was excluded due to its demonstrably low probability of yielding high CSNR. It was statistically informative but perceptually weak.

\subsubsection{Step 3: Optimization in the Green-Cyan Region}
The most critical decision involved selecting the third band from the spectrally crowded 470-525 nm region. The initial JMIM candidates, 477 nm and 520 nm, both showed high CSNR potential. However, the integrated analysis in Fig.~\ref{fig1} revealed they were situated in a zone of moderate-to-high inter-band correlation, making them spectrally redundant.

This prompted a search for a better trade-off. The 497 nm emerged as a superior choice. As shown in Fig.~\ref{fig1}, it lies in a blue region of low inter-band correlation while still maintaining a high CSNR probability. Although not in the top-five JMIM list, its proximity to the information-rich 477 nm band, combined with its superior performance on the other two criteria, made it the optimal compromise. This decision exemplifies the strength of our holistic approach over relying on a single metric.

This principled refinement process resulted in the final optimized band set: 497 nm, 607 nm, and 895 nm, which are the channel numbers 12, 40, and 114 in the H-City dataset, respectively. This trio represents the ideal balance of maximal statistical information, minimal redundancy, and high perceptual quality, as identified by our framework for the H-City dataset.

\begin{figure*}[hbt!]
\centering
{\includegraphics[width=0.999\textwidth]{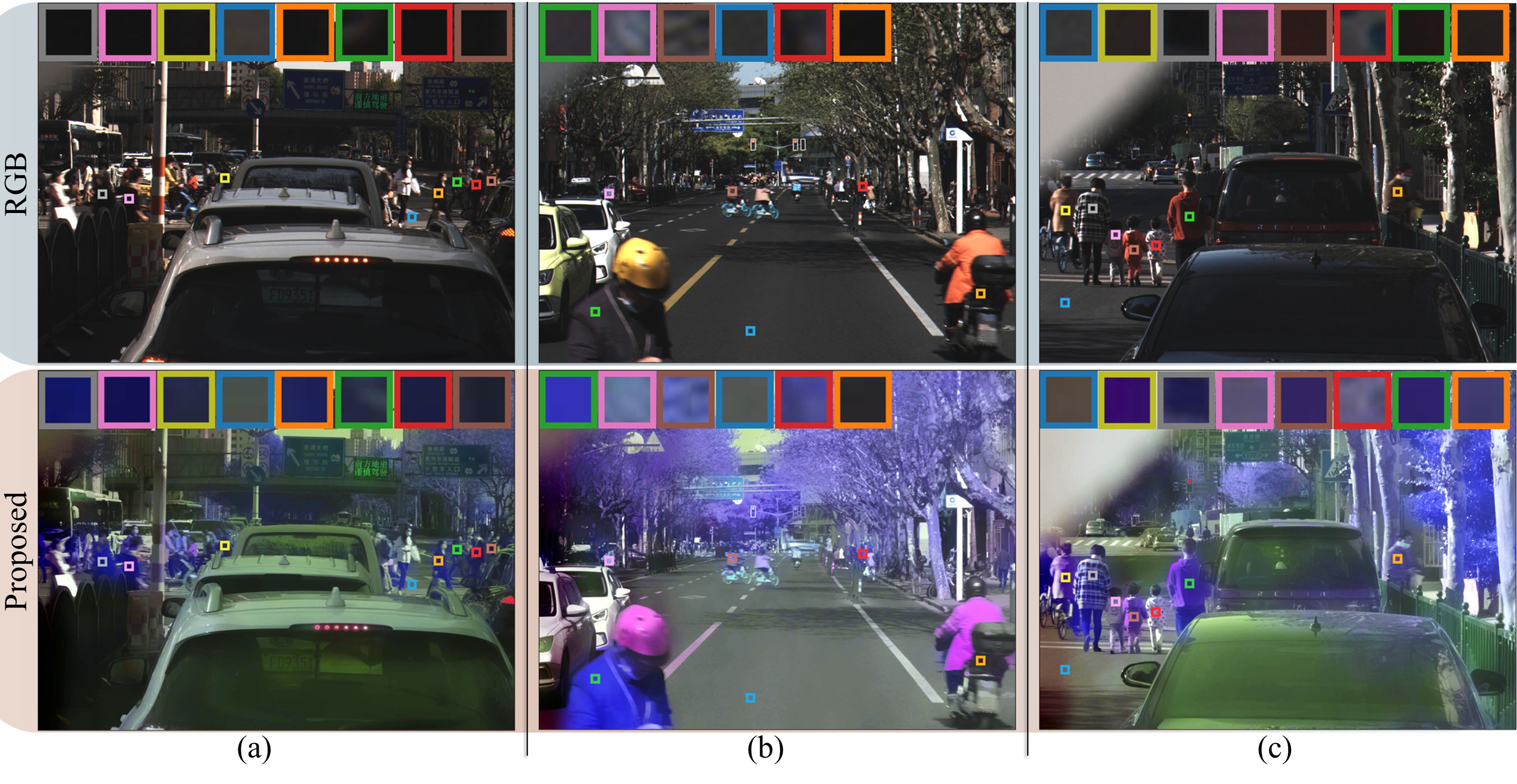}}
\caption{
Enhanced VRU discriminability using selected HSI bands compared to standard RGB. Each column (a–c) illustrates a challenging driving scenario containing multiple VRUs that exhibit potential metameric confusion with the asphalt background in the standard RGB image and the corresponding pseudo-color HSI composites (proposed approach). Distinct VRU patches are outlined with colored bounding boxes, consistently applied to the same spatial locations in RGB and HSI images. Asphalt background patches are marked with blue bounding boxes. Enlarged views of these patches are presented above their respective images.}
\label{fig_eval_visual_vru_first_part} 
\end{figure*}

\subsection{PSEUDO-COLOR COMPOSITE RECONSTRUCTION}

To create a visually intuitive representation for analysis, we reconstructed pseudo-color composites from the selected HSI bands. This process was designed to enhance salient spectral characteristics not visible in standard RGB images. The integrated response around 607 nm was mapped to the red channel and 497 nm to the green channel, analogous to their positions in the visible spectrum. Critically, the NIR band at 895 nm was mapped to the blue channel to translate a non-visible spectral advantage into a visible feature. Many materials used in VRU clothing exhibit high reflectance in the NIR region while appearing dark and indistinct from the background under certain lighting conditions~\cite{driggers2013good}. By mapping this NIR response to the blue channel, we create an artificial blue coloring for such objects, making them separable from the background in the pseudo-color composite. 

We integrated spectral data from seven adjacent bands on either side of each selected central wavelength ($\pm$27nm), totaling 15 bands per composite channel. This integration produces a broader spectral response around each central wavelength, thereby improving image clarity and promoting consistent brightness under varied illumination conditions. Finally, we applied standard gamma correction and white balance to optimize the composites for visual display.

These pseudo-color HSI composites provide a visually interpretable representation that not only preserves the inherent spatial structure of the scene but also harnesses the distinct spectral information captured by the selected HSI bands. This approach enables direct visual comparison with standard RGB images and ensures that the same patch-based quantitative analysis can be equitably applied to both image modalities.

\subsection{EVALUATION FRAMEWORK}
The evaluation of our selected HSI band set was structured to validate its effectiveness for ADAS/AD tasks. Our framework is built on two primary objectives and a robust set of comparison metrics.

\subsubsection{Evaluation Objectives}

Enhancing the separability of VRU is the primary objective of our band selection strategy. This is achieved by resolving metameric confusion and improving both the perceptual and quantitative distinctions of VRU from the background, addressing a key safety challenge for ADAS and AD.

The secondary objective is to achieve general scene legibility. We assess whether our band selection maintains or provides comparable differentiation for other vital road elements, particularly traffic lights and road markings, to guarantee broad utility.

\subsubsection{Comparison Metrics}
We employed both qualitative and quantitative methods to compare our pseudo-color HSI composites with standard RGB images of the same scenes. For qualitative assessment, corresponding RGB and HSI composites are presented side by side. Regions of interest are highlighted with identically colored bounding boxes for direct comparison, with enlarged views provided for detailed inspection.

To complement visual analysis, we applied a suite of quantitative metrics categorized into class dissimilarity metrics and visual perception metrics. The class dissimilarity metrics quantify the mathematical separability between the material patches $A$ and $B$ in feature space. Let $\mu_A$ and $\mu_B$ be the mean feature vectors, such as RGB triplets or 3-band HSI vectors. Detailed definitions of dissimilarity metrics are presented in the following.

\begin{enumerate}
    \item Euclidean distance ($D_2$) measures the straight-line distance between two vectors. A larger distance indicates greater dissimilarity. It is calculated as:
    $$
    D_2(\mu_A,\mu_B) = \sqrt{(\mu_A-\mu_B)^T(\mu_A-\mu_B)}.
    $$
    \item Spectral Angle Mapper (SAM) calculates the angle between two spectral vectors, making it robust to illumination variations. A larger angle signifies greater spectral difference. It is defined as: 
    $$
    \theta = cos^{-1} (\frac{\mu_A^T\mu_B}{|\mu_A||\mu_B|}).
    $$
    \item Multivariate $T^2$-test determines if two groups of vectors have significantly different means. A larger $T^2$ value indicates greater class separation. For all evaluated patches, the statistic $T^2$ yielded p-values lower than $0.001$, confirming a significant difference in the information captured. We therefore use the $T^2$ as a measure of the magnitude of this differentiation.
\end{enumerate}

Additionally, we employed the CIE $\Delta E$ color difference metric to quantify the perceptual difference between two colors in the CIELAB space, which is designed to be perceptually uniform. It is essential for assessing the risk of metameric confusion. A value of $\Delta E > 1.0$ is generally considered a noticeable difference to the human eye~\cite{choudhury1996evaluation}. A higher $\Delta E$ value indicates that two patches are more visually distinct. It is calculated as:
    $$
    \Delta E = \sqrt{(L_B - L_A)^2 + (a_B - a_A)^2 + (b_B - b_A)^2},    
    $$
where $(L_A, a_A, b_A)$ and $(L_B, a_B, b_B)$ are the CIELAB values of the patches $A$ and $B$. 

\begin{table*}[h!]
\centering
\caption{Metrics for VRU Separability: HSI Composite against RGB}
\label{tab:comparison_metrics_vru}
\resizebox{0.99\textwidth}{!}{%

\begin{tabular}{c|c|rr|rr|rr|rr}
\hline
\rule{0pt}{14pt}
\textbf{} & 
\textbf{Foreground} & \multicolumn{6}{c|}{\textbf{Dissimilarity Metrics}} & \multicolumn{2}{c}{\textbf{Perception Metric}}\\
\cline{3-10}
\textbf{Label} & \textbf{vs.} & 
\multicolumn{2}{c|}{\textbf{$D_2$}} & 
\multicolumn{2}{c|}{\textbf{SAM (in radians)}} & 
\multicolumn{2}{c|}{\textbf{$T^2$ (in $10^3$)}} & 
\multicolumn{2}{c}{\textbf{$\Delta E$}} \\
\cline{3-10}
 & \textbf{Asphalt} & 
\textbf{RGB} & \textbf{HSI Composite} & 
\textbf{RGB} & \textbf{HSI Composite} & 
\textbf{RGB} & \textbf{HSI Composite} & 
\textbf{RGB} & \textbf{HSI Composite} \\
\hline
(a) & Cloth 1 & 59.20 & \textbf{64.65} & 0.05 & \textbf{0.52} & 127.96 & \textbf{241.24} & 16.73 & \textbf{47.14}\\
    & Cloth 2 & \textbf{49.33} & 47.11 & 0.02 & \textbf{0.26} & 8.12 & \textbf{20.97} & 13.88 & \textbf{24.80}\\
    & Cloth 3 & 62.76 & \textbf{67.00} & 0.05 & \textbf{0.41} & 234.77 & \textbf{452.19} & 17.71 & \textbf{35.41} \\
    & Cloth 4 & 57.58 & \textbf{62.91} & 0.04 & \textbf{0.22} & 34.10 & \textbf{38.55} & 16.17 & \textbf{24.23} \\
    & Cloth 5 & 62.66 & \textbf{83.32} & 0.05 & \textbf{0.69} & 251.51 & \textbf{994.13} & 17.707 & \textbf{57.27} \\
    & Cloth 6 & 63.25 & \textbf{83.82} & 0.06 & \textbf{0.70} & 246.13 & \textbf{774.15} & 17.79 & \textbf{63.23} \\
    & Cloth 7 & \textbf{57.78} & 53.56 & 0.04 & \textbf{0.30} & 129.32 & \textbf{140.21} & 16.37 & \textbf{27.95} \\
\hline
(b) & Motorbike & 47.00 & \textbf{73.75} & 0.01 & \textbf{0.07} & 72.74 & \textbf{300.12} & 13.15 & \textbf{21.24} \\
    & Cloth 1 & 6.09 & \textbf{112.20} & 0.06 & \textbf{0.60} & 17.52 & \textbf{8821.22} & 5.21 & \textbf{86.24} \\
    & Cloth 2 & 13.16 & \textbf{58.97} & 0.03 & \textbf{0.35} & 1.37 & \textbf{67.95} & 6.12 & \textbf{46.49} \\
    & Cloth 3 & 19.18 & \textbf{87.55} & 0.02 & \textbf{0.31} & 0.90 & \textbf{34.93} & 6.67 & \textbf{47.08} \\  
    & Cloth 4 & 63.83 & \textbf{78.16} & 0.08 & \textbf{0.20} & 4.74 & \textbf{49.10} & 17.23 & \textbf{28.59} \\  
\hline
(c) & Cloth 1 & 35.85 & \textbf{54.51} & 0.10 & \textbf{0.40} & 71.81 & \textbf{1950.69} & 10.10 & \textbf{39.23} \\  
    & Cloth 2 & 49.43 & \textbf{70.96} & 0.14 & \textbf{0.58} & 155.14 & \textbf{3938.00} & 14.41 & \textbf{56.61} \\  
    & Cloth 3 & 4.61 & \textbf{70.03} & 0.04 & \textbf{0.26} & 0.53 & \textbf{68.13} & 5.79 & \textbf{29.92} \\  
    & Cloth 4 & 39.65 & \textbf{60.04} & 0.25 & \textbf{0.49} & 122.50 & \textbf{900.12} & 13.99 & \textbf{48.03} \\  
    & Cloth 5 & 7.42 & \textbf{53.39} & 0.07 & \textbf{0.25} & 12.30 & \textbf{144.99} & 5.09 & \textbf{27.06} \\ 
    & Cloth 6 & 44.70 & \textbf{57.86} & 0.02 & \textbf{0.47} & 8.41 & \textbf{373.68} & 12.50 & \textbf{39.14} \\ 
    & Cloth 7 & 31.15 & \textbf{78.96} & 0.10 & \textbf{0.65} & 39.82 & \textbf{810.82} & 9.05 & \textbf{64.99} \\ 
\hline
\multicolumn{2}{c|}{\textbf{Average}} & 40.77 & \textbf{69.41} & 0.06 & \textbf{0.41} & 81.04 & \textbf{1059.01} & 12.37 & \textbf{42.88} \\
\hline
\multicolumn{2}{c|}{\textbf{Improvement (\%)}} & \multicolumn{2}{c|}{\textbf{70.24}} & \multicolumn{2}{c|}{\textbf{528.46}} & \multicolumn{2}{c|}{\textbf{1206.83}} & \multicolumn{2}{c}{\textbf{246.62}} \\
\hline
\end{tabular}
}
\end{table*}

\subsection{RESULTS: ENHANCED VRU SEPARABILITY}
The primary goal of our HSI band selection is to improve the perception and differentiation of VRU, especially in scenarios where they are metameric with the background in RGB images. The results demonstrate a profound improvement, substantiated by visual (Fig.~\ref{fig_eval_visual_vru_first_part}) and quantitative evidence (Table~\ref{tab:comparison_metrics_vru}).

\begin{figure*}
\centering
{\includegraphics[width=0.999\textwidth]{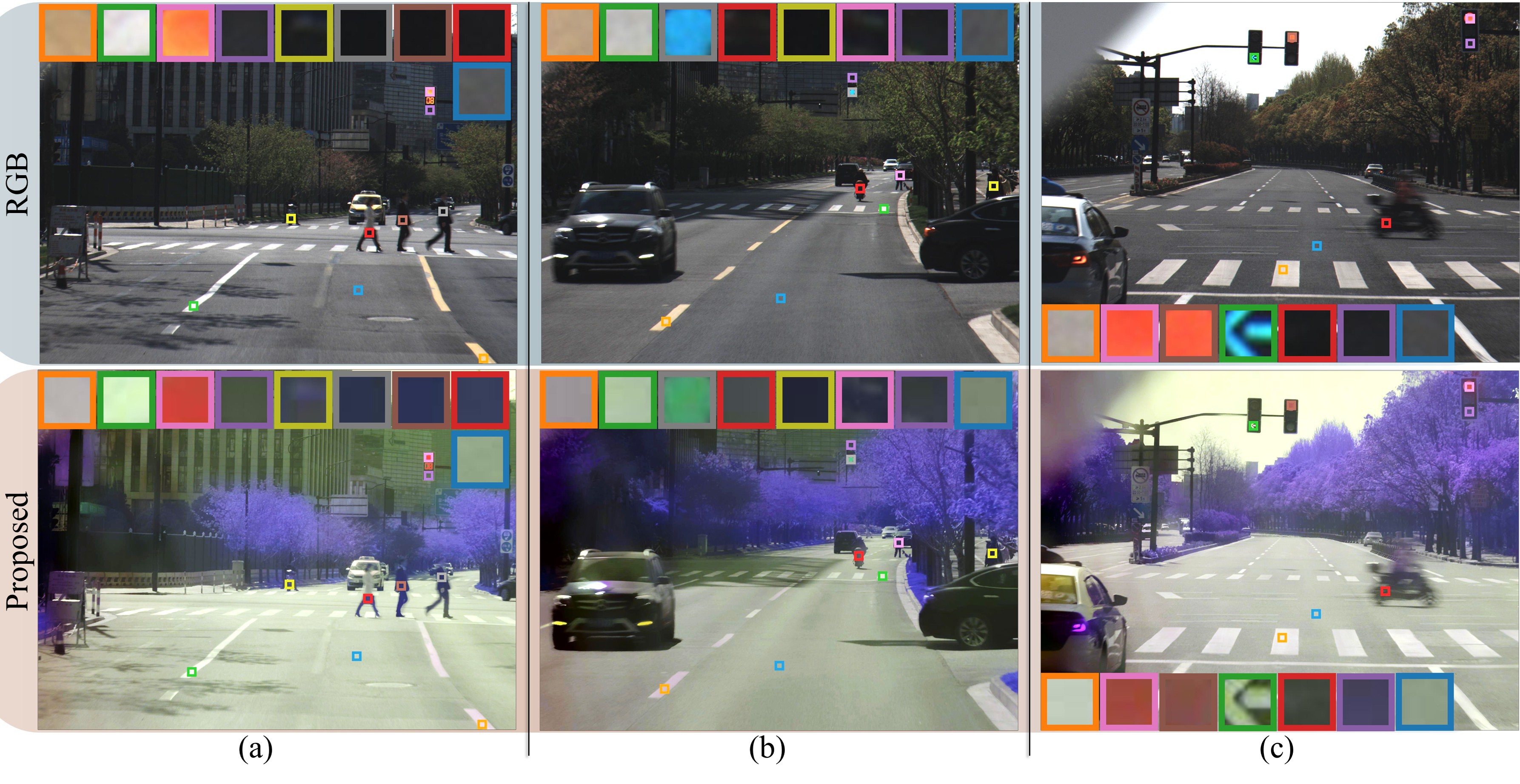}}
\caption{Visual comparison of selected HSI bands and standard RGB for diverse driving scene elements. Three distinct scenarios (a-c) compare standard RGB images with their corresponding pseudo-color HSI composites (proposed approach). Patches encompassing traffic lights (red, green), road markings (white, yellow), VRU, and asphalt are highlighted with colored bounding boxes, consistently applied across image pairs. Enlarged views are provided for each patch, respectively.}
\label{fig_eval_visual_other_first}
\end{figure*}

% \begin{figure*}
% \ContinuedFloat
% \centering

% % \caption{.}
% \label{fig_eval_visual_other_total}
% \end{figure*}

As illustrated in Fig.~\ref{fig_eval_visual_vru_first_part}, driving scenarios in which VRU appears nearly indistinguishable from the asphalt background in RGB are resolved in our HSI composites. The enlarged patches from the HSI images reveal a distinct and obvious color difference where the RGB patches show minimal distinction, confirming a clear visual enhancement of VRU perception.
Quantitatively, the HSI composites consistently and significantly outperformed RGB images across all metrics in distinguishing VRUs from the asphalt background. As shown in Table~\ref{tab:comparison_metrics_vru}, the improvements are substantial. Specifically, our HSI composites yielded increases over RGB by 70.24\% in Euclidean distance, 528.46\% in SAM, 1206.83\% in $T^2$, and 246.62\% in color difference. These metrics confirm that our pseudo-color HSI composites dramatically reduce metameric confusion for VRU, addressing a critical failure point of standard RGB sensors.

\subsection{RESULTS: Generalizability and Performance on Common Road Elements}
To assess the broader applicability of the selected HSI bands, we evaluated their performance in discerning other objects important for AD and ADAS, particularly traffic lights and road markings, which are typically well-defined in RGB images. The objective here was to ensure their separability was not compromised. 

The visual and quantitative data for this analysis are presented in Fig.~\ref{fig_eval_visual_other_first} and Table~\ref{tab:comparison_metrics_others}, reaffirming the primary strength of our band selection. They show that our HSI composites again demonstrate significantly enhanced separability for VRU against the background when compared to standard RGB. This consistent outperformance in varied scenes underscores the robustness of our approach for its safety objective.

The main purpose of this evaluation, however, was to verify that this targeted enhancement did not compromise the visibility of other common objects, particularly traffic lights and road markings. For these high-contrast elements, which are already clearly perceptible in standard RGB, the goal was to ensure the HSI composites maintain their separability. The quantitative analysis in Table~\ref{tab:comparison_metrics_others} confirms this. Although standard RGB may yield higher relative metric scores for these already high-contrast objects, the absolute metric values for the HSI composites demonstrate excellent discriminability. For instance, $D_2$ and $T^2$ values for traffic light and road marking pairs were consistently high, as were the perceptual $\Delta E$ values, far exceeding the just-noticeable-difference threshold ($\Delta E > 1.0$) and indicating no risk of metameric confusion.

This demonstrates that although our method is optimized for the challenging VRU problem, it maintains robust performance for other essential road scene components, ensuring no degradation in overall situational awareness.

\begin{table*}[htbp]
\centering
\caption{Metrics for Common Road Elements: HSI Composite against RGB}
\label{tab:comparison_metrics_others}
\resizebox{0.99\textwidth}{!}{%

\begin{tabular}{c|ll|rr|rr|rr|rr}
\hline
\rule{0pt}{14pt}
\textbf{} & 
\multicolumn{2}{c|}{\textbf{}} & 
\multicolumn{6}{c|}{\textbf{Dissimilarity Metrics}} &
\multicolumn{2}{c}{\textbf{Perception Metric}} \\
\cline{4-11}
\textbf{Label} & 
\multicolumn{2}{c|}{\textbf{Patches}} &
\multicolumn{2}{c|}{\textbf{$D_2$}} & 
\multicolumn{2}{c|}{\textbf{SAM (in radians)}} & 
\multicolumn{2}{c|}{\textbf{$T^2$ (in $10^3$)}} & 
\multicolumn{2}{c}{\textbf{$\Delta E$}} \\
\cline{2-11}%{3-12}
 
 & \textbf{Foreground} & \textbf{Background} &
\textbf{RGB} & \textbf{HSI Composite} & 
\textbf{RGB} & \textbf{HSI Composite} & 
\textbf{RGB} & \textbf{HSI Composite} & 
\textbf{RGB} & \textbf{HSI Composite} \\
\hline
(a) & Cloth 1 & Asphalt & 98.09 & \textbf{146.09} & 0.03 & \textbf{0.28} & 172.75 & \textbf{1009.74} & 25.83 & \textbf{54.02} \\
    & Cloth 2 & Asphalt & 107.16 & \textbf{163.77} & 0.03 & \textbf{0.35} & 219.85 & \textbf{1299.81} & 28.38 & \textbf{60.32} \\
    & Cloth 3 & Asphalt & 107.85 & \textbf{166.29} & 0.05 & \textbf{0.29} & 225.01 & \textbf{985.01} & 28.68 & \textbf{56.08} \\
    & Motorbike & Asphalt & 93.61 & \textbf{135.91} & 0.05 & \textbf{0.17} & 42.62 & \textbf{266.09} & 24.62 & \textbf{45.55} \\
    \cdashline{2-11}
    & Yellow Line & Asphalt & \textbf{130.37} & 57.79 & \textbf{0.19} & 0.09 & 78.78 & \textbf{368.89} & \textbf{40.19} & 20.36 \\
    & White Line & Asphalt & \textbf{236.44} & 116.26 & \textbf{0.02} & 0.01 & \textbf{25.76} & 24.04 & \textbf{52.53} & 27.20 \\
    & Yellow Line & White Line & \textbf{125.23} & 66.97 & \textbf{0.17} & 0.09 & 50.20 & \textbf{63.95} & 33.23 & \textbf{30.50} \\
    & Traffic Light & Traffic Light Off & \textbf{232.13} & 130.82 & 0.47 & \textbf{0.57} & \textbf{105.00} & 40.85 & \textbf{85.02} & 68.20 \\
    
\hline
% (b) & Cloth 1 & Asphalt & 154.43 & \textbf{188.68} & 0.04 & \textbf{0.28} & 222.20 & \textbf{1474.27} & 39.72 & \textbf{60.43} \\ 
%     & Cloth 2 & Asphalt & 141.99 & \textbf{163.93} & 0.04 & \textbf{0.20} & 129.77 & \textbf{208.13} & 36.23 & \textbf{50.94} \\
%     & Motorbike 1 & Asphalt & 145.48 & \textbf{173.06} & 0.04 & \textbf{0.27} & 117.59 & \textbf{238.47} & 37.18 & \textbf{57.54} \\
%     & Motorbike 2 & Asphalt & 92.98 & \textbf{110.00} & \textbf{0.11} & 0.04 & \textbf{26.22} & 16.80 & 24.07 & \textbf{28.02} \\ 
%     \cdashline{2-11}
%     & White Line & Asphalt & \textbf{127.32} & 45.73 & 0.00 & \textbf{0.01} & \textbf{64.67} & 34.11 & \textbf{28.20} & 10.76 \\
%     & Traffic Light 1 & Traffic Light Off & \textbf{263.50} & 165.74 & 0.42 & \textbf{0.51} & 287.47 & \textbf{3120.63} & 73.56 & \textbf{106.29} \\
%     & Traffic Light 2 & Traffic Light Off & \textbf{62.73} & 62.67 & \textbf{0.33} & 0.25 & 1.730 & \textbf{17.63} & 25.26 & \textbf{34.07} \\
% \hline
(b) & Cloth 1 & Asphalt & 83.85 & \textbf{125.60} & 0.04 & \textbf{0.21} & 33.150 & \textbf{49.76} & 22.33 & \textbf{41.45} \\ 
    & Cloth 2 & Asphalt & 102.94 & \textbf{146.80} & 0.03 & \textbf{0.26} & 448.11 & \textbf{2301.42} & 27.53 & \textbf{46.87} \\    
    & Motorbike & Asphalt & 87.96 & \textbf{88.28} & 0.02 & \textbf{0.07} & 91.823 & \textbf{190.50} & 23.33 & \textbf{26.22} \\  
    \cdashline{2-11}
    & Yellow Line & Asphalt & \textbf{144.82} & 50.96 & \textbf{0.16} & 0.08 & \textbf{241.11} & 123.02 & \textbf{40.33} & 17.66 \\  
    & White Line & Asphalt & \textbf{178.01} & 88.20 & \textbf{0.03} & 0.01 & 31.55 & \textbf{55.03} & \textbf{41.46} & 21.59 \\  
    & Yellow Line & White Line & \textbf{53.88} & 45.67 & \textbf{0.13} & 0.09 & 29.87 & \textbf{53.26} & 20.10 & \textbf{23.68} \\
    & Traffic Light  & Traffic Light Off & \textbf{203.18} & 90.88 & \textbf{0.46} & 0.23 & 4.76 & \textbf{5.402} & \textbf{59.50} & 44.92 \\ 
\hline
(c) & Motorbike & Asphalt & 75.13 & \textbf{138.53} & 0.01 & \textbf{0.02} & 139.60 & \textbf{425.00} & 20.05 & \textbf{34.68} \\ 
\cdashline{2-11}
    & White Line & Asphalt & \textbf{162.52} & 87.92 & \textbf{0.05} & 0.03 & 303.98 & \textbf{392.19} & \textbf{38.55} & 18.46 \\  
    & Traffic Light Green & Traffic Light Off & \textbf{128.90} & 87.05 & \textbf{0.38} & 0.26 & 1.35 & \textbf{5.82} & 42.95 & \textbf{46.94} \\   
    & Traffic Light Red1 & Traffic Light Off & \textbf{224.10} & 66.12 & \textbf{0.57} & 0.38 & 32.54 & \textbf{59.19} & \textbf{88.17} & 33.42 \\  
    & Traffic Light Red2 & Traffic Light Off & \textbf{229.02} & 96.74 & \textbf{0.60} & 0.53 & \textbf{4891.67} & 222.20 & \textbf{92.36} & 50.30 \\ 
    & Traffic Light Green & Traffic Light Red1 & \textbf{222.36} & 66.93 & \textbf{0.95} & 0.27 & \textbf{30.89} & 16.20 & \textbf{102.12} & 40.24 \\
    & Traffic Light Green & Traffic Light Red2 & \textbf{230.04} & 87.70 & \textbf{0.97} & 0.42 & \textbf{420.44} & 133.62 & \textbf{106.80} & 53.72 \\
        
\hline
\end{tabular}
}
\end{table*}

\section{Discussion}
\label{discussion}
The core finding of this research is that our HSI band selection strategy significantly enhances VRU perception by resolving metameric ambiguities that confound standard RGB sensors, a finding confirmed by superior dissimilarity ($D_2$, SAM, and $T^2$), and perception ($\Delta E$) metrics. A key insight is the pronounced effectiveness of the 895 nm NIR wavelength. Mapping its high reflectance to the blue channel creates a strong, artificial color contrast that makes otherwise indistinct VRU visually salient. This work provides evidence that a minimal set of strategically chosen bands, including one in the NIR spectrum, can overcome a critical failure mode in automotive vision.

Methodologically, this study introduces CSNR as a novel, perceptually-relevant metric for HSI band selection in AD. Combined with information theory (JMIM) and correlation analysis, our holistic framework successfully identified bands that were both statistically informative and visually effective. The primary practical outcome of this 3-band solution is a $97.6\%$ reduction in data volume compared to the full HSI cube. This significant data compression is a critical step toward making HSI processing computationally feasible for on-vehicle, real-time applications.

We acknowledge that our classifier-independent approach does not capture interactions with specific downstream classifier architectures. Therefore, future work should focus on integrating our selected bands with state-of-the-art object detection models to quantify the end-to-end performance benefits. Validating the generalizability of the framework across new HSI datasets, as they become available, is also a critical next step.

\section{CONCLUSION}
\label{conclusion}
This research establishes that a strategically selected three-band HSI system (497 nm, 607 nm, 895 nm) can resolve critical metameric ambiguities and significantly enhance VRU safety beyond the capabilities of standard RGB sensors. Our findings prove that this targeted approach successfully improves the perception of the VRU without compromising the visibility of other essential road elements.

The introduction of CSNR as a novel selection metric provides a more perceptually grounded framework for future HSI research. Critically, the $97.6\%$ data reduction achieved by our framework makes the use of HSI technology more feasible for real-world, on-vehicle computation. Although future work will involve deeper integration with machine learning classifiers and validation on broader datasets, this study provides a robust methodology and a practical pathway for leveraging strategic HSI bands to build safer, more reliable automotive perception systems.

% \section*{REFERENCES}
\bibliographystyle{IEEEtran}
\bibliography{refs}

\end{document}